\documentclass[sigconf]{acmart}
\usepackage{multirow}
\usepackage{multicol}
\usepackage{caption}
\usepackage{xcolor} 
\usepackage{color} 
\usepackage{verbatim} 
\newcommand{\tabincell}[2]{\begin{tabular}{@{}#1@{}}#2\end{tabular}}
\AtBeginDocument{%
  \providecommand\BibTeX{{%
    \normalfont B\kern-0.5em{\scshape i\kern-0.25em b}\kern-0.8em\TeX}}}

\setcopyright{acmcopyright}
\copyrightyear{2018}
\acmYear{2018}
\acmDOI{10.1145/1122445.1122456}

\acmConference[Woodstock '18]{Woodstock '18: ACM Symposium on Neural
  Gaze Detection}{June 03--05, 2018}{Woodstock, NY}
\acmBooktitle{Woodstock '18: ACM Symposium on Neural Gaze Detection,
  June 03--05, 2018, Woodstock, NY}
\acmPrice{15.00}
\acmISBN{978-1-4503-XXXX-X/18/06}

\author{Qinyan Dai\textsuperscript{1 $\dagger$}, Juncheng Li\textsuperscript{1}, Qiaosi Yi\textsuperscript{1}, Faming Fang\textsuperscript{1 *}, Guixu Zhang\textsuperscript{1}\\ \small \textsuperscript{1}School of Computer Science and Technology, East China Normal University, Shanghai, China\\ \small \textsuperscript{$\dagger$}51194506008@stu.ecnu.edu.cn \,\,\,  \textsuperscript{*}fmfang@cs.ecnu.edu.cn}

\begin{document}

\title{Feedback Network for Mutually Boosted Stereo Image Super-Resolution and Disparity Estimation}

\begin{abstract}
Under stereo settings, the problem of image super-resolution (SR) and disparity estimation are interrelated that the result of each problem could help to solve the other. The effective exploitation of correspondence between different views facilitates the SR performance, while the high-resolution (HR) features with richer details benefit the correspondence estimation. According to this motivation, we propose a Stereo Super-Resolution and Disparity Estimation Feedback Network (SSRDE-FNet), which simultaneously handles the stereo image super-resolution and disparity estimation in a unified framework and interact them with each other to further improve their performance. Specifically, the SSRDE-FNet is composed of two dual recursive sub-networks for left and right views. Besides the cross-view information exploitation in the low-resolution (LR) space, HR representations produced by the SR process are utilized to perform HR disparity estimation with higher accuracy, through which the HR features can be aggregated to generate a finer SR result. Afterward, the proposed HR Disparity Information Feedback (HRDIF) mechanism delivers information carried by HR disparity back to previous layers to further refine the SR image reconstruction. Extensive experiments demonstrate the effectiveness and advancement of SSRDE-FNet.
\end{abstract}



\maketitle

\section{Introduction}
With the development of dual cameras, stereo images have shown greater impact in many applications, such as smartphones, drones, and autonomous vehicles. However, the stereo images often suffer from resolution degradation in practice. Therefore, a technology that can restore the high-resolution (HR) left and right views in a 3D scene is essential. In the binocular system, parallax effects between the low resolution (LR) images cause a sub-pixel shift between them. Therefore, making full use of cross-view information can help reconstruct high-quality SR images since one view may have additional information relative to the other.

Recently, several deep learning based methods have been proposed to capture cross-view information by modeling the disparity. For example, ~\cite{Wang2019LearningPA, Wang2020ParallaxAF, Ying2020ASA, Song2020StereoscopicIS, BSSRnet, Wang2020SymmetricPA} leverage the parallax attention module (PAM) proposed by Wang et al.~\cite{Wang2019LearningPA, Wang2020ParallaxAF} to search for correspondences along the horizontal epipolar line without parallax limit; In ~\cite{Yan2020DisparityAwareDA}, a pre-trained disparity network~\cite{Khamis2018StereoNetGH} was used to deploy the disparity prior into image reconstruction. Although continuous improvements have been achieved in stereo image SR, the utilization of cross-view information is still insufficient and less effective.

In fact, under stereo settings, disparity estimation and image SR are interrelated that the result of each problem could help to solve the other one, and each task benefits from the gradual improvement over the other task. However, previous methods have not explored this mutually boosted property. Moreover, all these methods exploit correspondent information only in the LR space, which usually does not provide enough accuracy in high-frequency regions due to the loss of fine-grained details in LR features. Thus, the positive additional information brought by these correspondences is still limited, hindering sufficient feature aggregation and further SR performance improvements. Thus, it is highly desirable to model disparity in a more powerful way and have a guidance mechanism that can fully interact between super-resolution and disparity estimation.

\begin{figure*}[t]
  \centering
  \includegraphics[width=16cm]{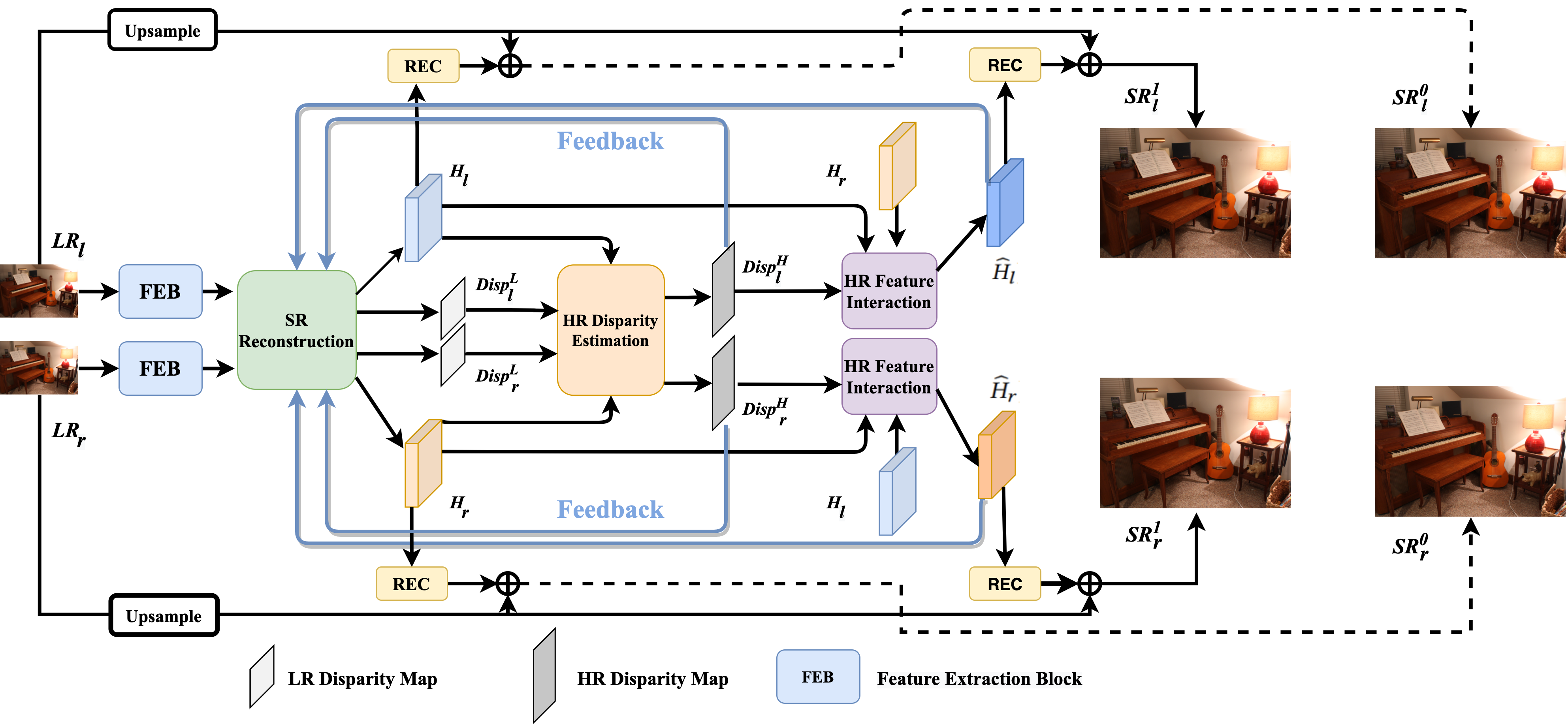}
  \begin{minipage}[c]{1\textwidth}
  \end{minipage}
  \caption{The architecture of SSRDE-FNet, which introduces the HR disparity information feedback mechanism.}
  \label{fig-2}
  \vspace{-12px}
\end{figure*}

To address the aforementioned problem, we propose a novel method that can handle stereo image super-resolution and HR disparity estimation in an end-to-end framework (Figure~\ref{fig-2}), interacting in a mutually boosted manner. We perform disparity estimation in the HR space to overcome the accuracy limitation of LR correspondence and better guide the stereo SR. To achieve this efficiently, we leverage the features from LR space and the reconstructed HR space to estimate disparity in a coarse-to-fine manner. In the framework, the guidance and interaction of super-resolution and disparity estimation are three-folds: (i). the coarse correspondence estimation in LR space benefits the cross-view information exploration for SR, initial SR results and HR features for both views are produced; (ii). the HR representations from (i) with richer details serve as finer features for HR disparity estimation, which reduces the search range of HR disparity for better accuracy and efficiency; (iii). The HR disparity can further benefit SR reconstruction. Specifically, we align the HR features of the two views using HR disparity maps and perform attention-driven feature aggregation to produce the enhanced HR features, upon which a finer SR result is generated. To achieve a more essential facilitation of HR disparity to stereo SR, we propose the HR Disparity Information Feedback (HRDIF) mechanism that feeds the enhanced HR features and the HR disparity back to previous layers for the refinement of low-level features in the SR process. In summary, the main contributions of this paper are as follows:

\begin{itemize}
	\item We propose a Stereo Super-Resolution and Disparity Estimation Feedback Network (SSRDE-FNet) that can simultaneously solve the stereo image super-resolution and disparity estimation in a unified framework. To the best of our knowledge, this is the first end-to-end network that can achieve the mutual boost of these two tasks.
	
	\item We propose a novel HR Disparity Information Feedback (HRDIF) mechanism for HR disparity and promote the quality of the SR image in an iterative manner.
	
	\item Extensive experiments illustrate that the proposed model can restore high-quality SR images, and the model achieves state-of-the-art results in the field of stereo image super-resolution.
\end{itemize}

\section{Related Works}
\subsection{Image Super-Resolution}
Image Super-Resolution aims to reconstruct a super-resolution (SR) image from its degraded low-resolution (LR) one, which is an extremely hot topic in the computer vision field. Since the pioneer work of Super-Resolution Convolutional Neural Network (SRCNN~\cite{Dong2014LearningAD}), learning-based methods have dominated the research of single image super-resolution (SISR). Methods like VDSR~\cite{Kim2016AccurateIS}, SRDenseNet~\cite{Tong2017ImageSU}, EDSR~\cite{Lim2017EnhancedDR}, MSRN~\cite{Li2018MultiscaleRN}, and RDN~\cite{Zhang2018ResidualDN} achieved excellent performance and greatly promoted the development of SISR. However, due to the lack of reference features, the development of SISR has encountered a bottleneck, and its performance is difficult to further improve. Therefore, stereo image super-resolution has received great attention in recent years since it has the available left and right view information. The critical challenge for enhancing spatial resolution from stereo images is how to register corresponding pixels with sub-pixel accuracy. Bhavsar et al.~\cite{Bhavsar2010ResolutionEI} argued that the twin problems of image SR and HR disparity estimation are intertwined under stereo settings. They formulate the two problems into one energy function, and minimize it by iteratively updating the HR image and disparity map. The following conventional methods~\cite{Park2012CombiningMS, Lee2013SimultaneousSO} usually follow this pipeline, however, these methods usually take a large amount of computation time. Recently, several deep learning-based stereo SR methods have emerged by using the parallax. For example, StereoSR~\cite{Jeon2018EnhancingTS} stacks stereo images with horizontal shift intervals to feed into the network to learn stereo correspondences. However, the maximum parallax that can be processed is fixed as 64. To explore correspondences without disparity limit, Wang et al.~\cite{Wang2019LearningPA, Wang2020ParallaxAF} proposed PASSRnet, with a parallax-attention module (PAM) that has a global receptive field along the epipolar line for global correspondence capturing. Ying et al.~\cite{Ying2020ASA} and Song et al.~\cite{Song2020StereoscopicIS} also made use of the PAM, while~\cite{Ying2020ASA} incorporated several PAMs to different stages of the pre-trained SISR networks to enhance the cross-view interaction. In iPASSR~\cite{Wang2020SymmetricPA}, a symmetric bi-directional PAM (biPAM) and an inline occlusion handling scheme are proposed to further improve SR performance. Besides the PAM based methods, Yan et al.~\cite{Yan2020DisparityAwareDA} uses a pre-trained disparity flow network to predict the disparity map based on the input stereo pair, and incorporates the disparity prior to better utilize the cross-view nature. Lei et al.~\cite{Lei2020DeepSI} builds up an interaction module-based stereo SR network (IMSSRnet), in which the interaction module is composed of a series of interaction units with a residual structure.

Above methods all explore the correspondence information between stereo images only in the LR space, limiting the positive effects provided by cross-view. Our work hunts for the mutual contributions between the stereo image SR and HR disparity estimation, leading to higher image quality and more accurate disparity, which is new in literature w.r.t learning-based method.

\subsection{Disparity Estimation}
Disparity estimation has been investigated to obtain correspondence between a stereo image pair~\cite{Scharstein2004ATA, Luo2016EfficientDL}, which can be utilized to capture long-range dependency for stereo SR. Existing end-to-end disparity estimation networks usually include cost volume computation, cost aggregation, and disparity prediction. 2D CNN based methods~\cite{Mayer2016ALD, Liang2018LearningFD, Xu2020AANetAA} generally adopt a correlation layer for 3D cost volume construction, while 3D CNN based methods~\cite{Kendall2017EndtoEndLO, Chang2018PyramidSM, Nie2019MultiLevelCU, Zhang2019GANetGA, Chabra2019StereoDRNetDR} mostly use direct feature concatenation to construct 4D cost volume and use 3D convolution for cost aggregation. However, learning matching costs from 4D cost volumes suffers from a high computational and memory burden. Apart from supervised methods, several unsupervised learning methods~\cite{Zhou2017UnsupervisedLO, Li2018OcclusionAS, Yang2018SegStereoES, Pilzer2020ProgressiveFF, Wang2020ParallaxAF} have been developed to avoid the use of costly ground truth depth annotations. Most relevantly, Wang et al.~\cite{Wang2020ParallaxAF} uses cascaded PAM to regress matching costs in a coarse-to-fine manner, getting rid of the limitation of fixed maximum disparity in cost volume techniques. However, as Gu at al.~\cite{Gu2020CascadeCV} pointed out, due to computational limitation, methods usually calculate matching cost at a lower resolution by the downsampled feature maps and rely on interpolation operations to generate HR disparity. Differently, they decompose the single cost volume into a cascade formulation of multiple stages for efficient HR stereo matching. Inspired by this, we achieve the HR disparity estimation in a coarse-to-fine manner.

\section{Method}
As shown in Figure~\ref{fig-2}, we develop a Stereo Super-Resolution and Disparity Estimation Feedback Network (SSRDE-FNet) in this paper. The goal of our method is to obtain SR images $SR_{l}$, $SR_{r}$ of both view and relevant HR disparity maps $D^{HR}_{l}$, $D^{HR}_{r}$, from LR stereo images input $LR_{l}$, $LR_{r}$, and interact the two tasks in a mutually boosted way. In this section, we first introduce the overall insights and network architecture in Sec.~\ref{network}. Then, we detail the novel proposed HR Disparity Information Feedback (HRDIF) mechanism in Sec.~\ref{feedback}. Finally, the loss functions are presented in Sec.~\ref{loss}.

\subsection{SSRDE-FNet}\label{network}
A key to improve stereo SR is utilizing disparity for sub-pixel information registration, and a key to disparity estimation accuracy is the resolution of input features. To let these two tasks make mutually effective contribution to each other, the modeling power of both tasks are important. Thus, we propose a Stereo Super-Resolution and Disparity Estimation Feedback Network (SSRDE-FNet). As shown in Figure.~\ref{fig-2}, SSRDE-FNet is essentially a recurrent network with the proposed HR Disparity Information Feedback (HRDIF) mechanism. In each iteration, two SR reconstruction steps are involved. The HR disparity is achieved in a coarse-to-fine way, coarse disparity is first estimated from LR features and the finer one is estimated from the reconstructed HR features. The advantages of this method are: (1) Stereo image SR can utilize cross-view information in multi-scales since both LR and HR correspondences can be obtained, leading to more sufficient feature aggregation; (2) The coarse-to-fine manner leads to a more compact and efficient network.

\begin{figure}[t]
  \centering
  \includegraphics[width=8.5cm]{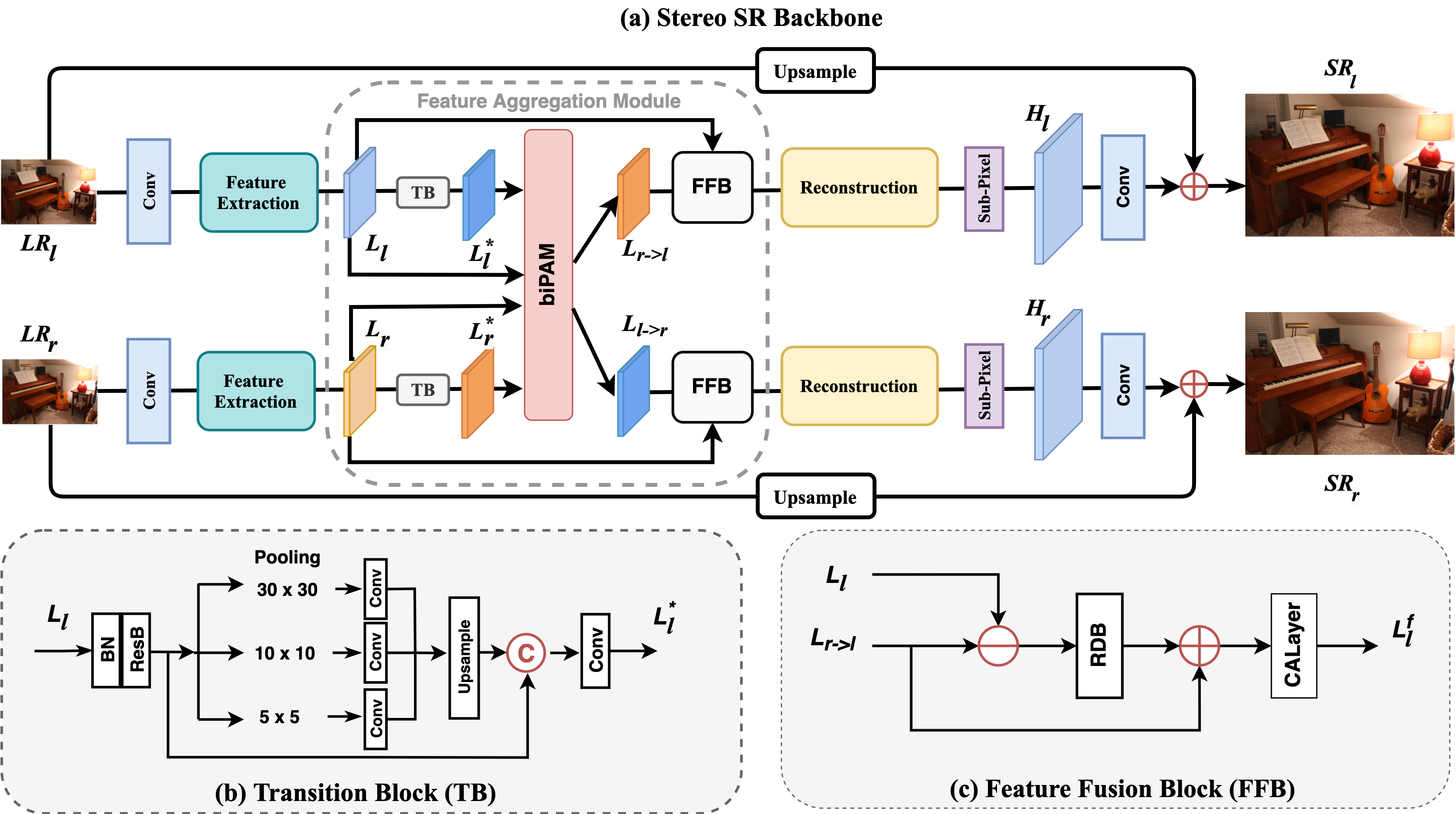}
  \begin{minipage}[c]{1\textwidth}
  \end{minipage}
  \setlength{\belowcaptionskip}{-0pt} 
  \setlength{\belowcaptionskip}{-10pt}
  \caption{The architecture of the proposed SR backbone.}
  \label{fig-3}
\end{figure}

\textbf{Stereo Image SR Backbone}
We develop a lightweight stereo SR network as shown in Figure~\ref{fig-3}(a), which leverages both intra-view and cross-view LR information for image reconstruction. Since hierarchical features have been demonstrated to be effective in both SISR~\cite{Li2018MultiscaleRN,Zhang2018ResidualDN} and disparity estimation~\cite{Kendall2017EndtoEndLO,Chang2018PyramidSM}, we are also committed to maximizing the use of hierarchical features in the model. Specifically, after a convolution layer that extracts shallow features, four RDBs~\cite{Zhang2018ResidualDN} are stacked to extract hierarchical features. Finally, we make full use of the features from all the RDBs by concatenating them and fusing them with a $1 \times 1$ convolution. Meanwhile, in order to alleviate the training conflict that may suffered by directly sharing features across different tasks~\cite{Sener2018MultiTaskLA} and explore more adaptive features for LR disparity estimation, a transition block is performed on $L_{l}$ and $L_{r}$, expressed as:
\begin{equation}\label{eq-1}
\small
\setlength{\belowcaptionskip}{-8pt}
 L^{*}_{l}=f_{TB}(L_{l}), L^{*}_{r}=f_{TB}(L_{r}).
\end{equation}
Among them, $L_{l}$ and $L_{r}$ denote the extracted features, $L^{*}_{l}$ and $L^{*}_{r}$ denote the transformed features, and $f_{TB}$ denotes the transition block (TB). As shown in Figure~\ref{fig-3}(b)), we apply a Spatial Pyramid Pooling (SPP) module in the TB for multi-scale feature extraction, which can further improve model performance.

Under LR space, we explore cross-view information by sampling disparity across the entire horizontal-range of a scene.
To achieve this, bi-directional parallax attention module (biPAM~\cite{Wang2020SymmetricPA}) is adopted. In this work, it serves as both self-attention LR feature registration and coarse disparity estimation for HR disparity initialization, thus its reliability is important. However, even with deep features, matching from unaries is far from reliable. To this end, 
we cascade $N$ biPAMs for matching cost aggregation. Therefore, the operation of the $i^{th}$ biPAM can be defined as:

\begin{equation}\label{eq-2}
\small
\setlength{\belowcaptionskip}{-8pt}
\begin{split}
 L^{'}_{l} =&f_{CONV}(L^{*,i-1}_{l}), L^{'}_{r}=f_{CONV}(L^{*,i-1}_{r}), \\
 &\mathbf{C}_{l\rightarrow r}^{i} =\mathbf{C}_{l\rightarrow r}^{i-1}+f_{Q}(L^{'}_{l})\otimes f_{K}(L^{'}_{r})^{T}, \\
 &\mathbf{C}_{r\rightarrow l}^{i} =\mathbf{C}_{r\rightarrow l}^{i-1}+f_{Q}(L^{'}_{r})\otimes f_{K}(L^{'}_{l})^{T}, \\
 & L^{*,i}_{l} =L^{*,i-1}_{l}+L^{'}_{l},
 L^{*,i}_{r}=L^{*,i-1}_{r}+L^{'}_{r},
\end{split}
\end{equation}

where $f_{CONV}$ denotes two $3 \times 3$ convolutions. $f_{Q}$ and $f_{K}$ are both $1 \times 1$ convolution. $\otimes$ is geometry-aware matrix multiplication, T is transposition operation that exchanges the last two dimensions of a matrix. Finally, the softmax is applied on $\mathbf{C}_{l\rightarrow r}^{N}$ and $\mathbf{C}_{r\rightarrow l}^{N}$ to generate parallax attention map $\mathbf{M}_{l\rightarrow r}^{N}$ and $\mathbf{M}_{r\rightarrow l}^{N}$. Therefore, the warped feature maps $L_{r\rightarrow l}, L_{l\rightarrow r}$ for sub-pixel registration are generated by the corresponding parallax attention map and inline occlusion inline occlusion handling~\cite{Wang2020SymmetricPA}. For each view, its own feature and the warped feature from the other view are then sent to the feature fusion module (FFM) for cross-view information aggregation. Instead of directly concatenate the two features, we build a residual based aggregation module (Fig.~\ref{fig-3}(c)). To allow the network to concentrate on more informative features that are complementary from cross-view, we first compute the residual between the two features, and then apply a RDB~\cite{Zhang2018ResidualDN} on the residual features, the output features are then added back to the view's own feature. Take the left view as example, the operation can be defined as:

\begin{equation}\label{eq-5}
\small
\setlength{\belowcaptionskip}{-8pt}
\begin{split}
 Res_{l}&=L_{r\rightarrow l}-L_{l}, \\
 L^{f}_{l}&=f_{CALayer}(f_{RDB}(Res_{l})+L_{l}), 
 \end{split}
\end{equation}
where $L^{f}_{l}$ denotes the fused features for left view and $f_{CALayer}$ denotes the channel attention layer. Such inter-residual projection allows the network to focus only on the distinct information between feature sources while bypassing the common knowledge, enabling a more discriminative feature aggregation compared with trivial adding or concatenating. Finally, the fused features $L^{f}_{l}, L^{f}_{r}$ go through the reconstruction module that has the same architecture with the feature extraction module, and a sub-pixel convolutional layer is applied to produce the HR feature $H_{l}, H_{r}$. Meanwhile, the SR images $SR^{0}_{l}, SR^{0}_{r}$ are reconstructed at this step by adding the corresponding bicubic upsampled LR images: 

\begin{equation}\label{eq-13}
\small
\setlength{\belowcaptionskip}{-8pt}
\begin{split}
 SR^{0}_{l}=f_{UP}(LR_{l})+f_{REC}(H_{l}), \\
 SR^{0}_{r}=f_{UP}(LR_{r})+f_{REC}(H_{r}). 
 \end{split}
\end{equation}

The main role of the two super-resolved images is to guarantee the effectiveness of the HR features $H_{l}, H_{r}$, which serve as important inputs to the subsequent HR disparity estimation module.

\begin{figure}
  \centering
  \includegraphics[width=8.5cm]{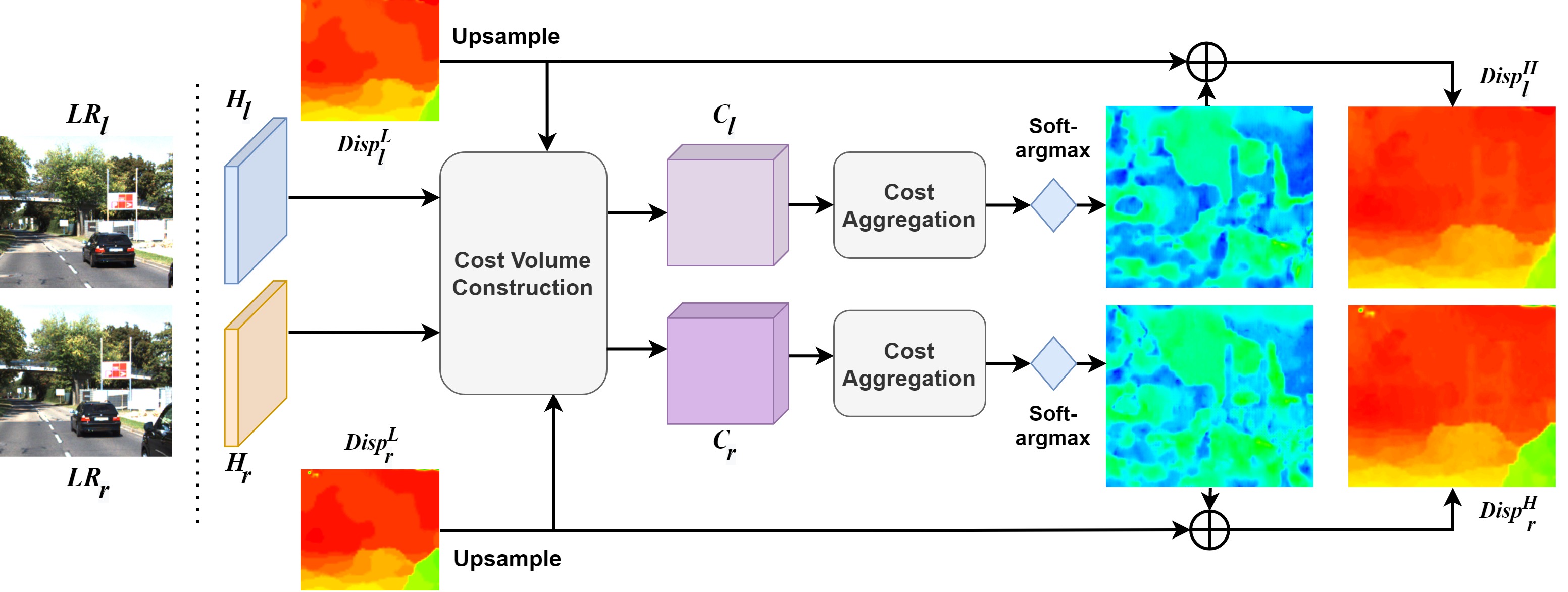}
  \begin{minipage}[c]{1\textwidth}
  \end{minipage}
  \caption{Illustration of HR disparity estimation module.}
  \label{fig-5}
  \vspace{-8pt}
\end{figure}

\begin{figure*}[t]
  \centering
  \includegraphics[width=12cm]{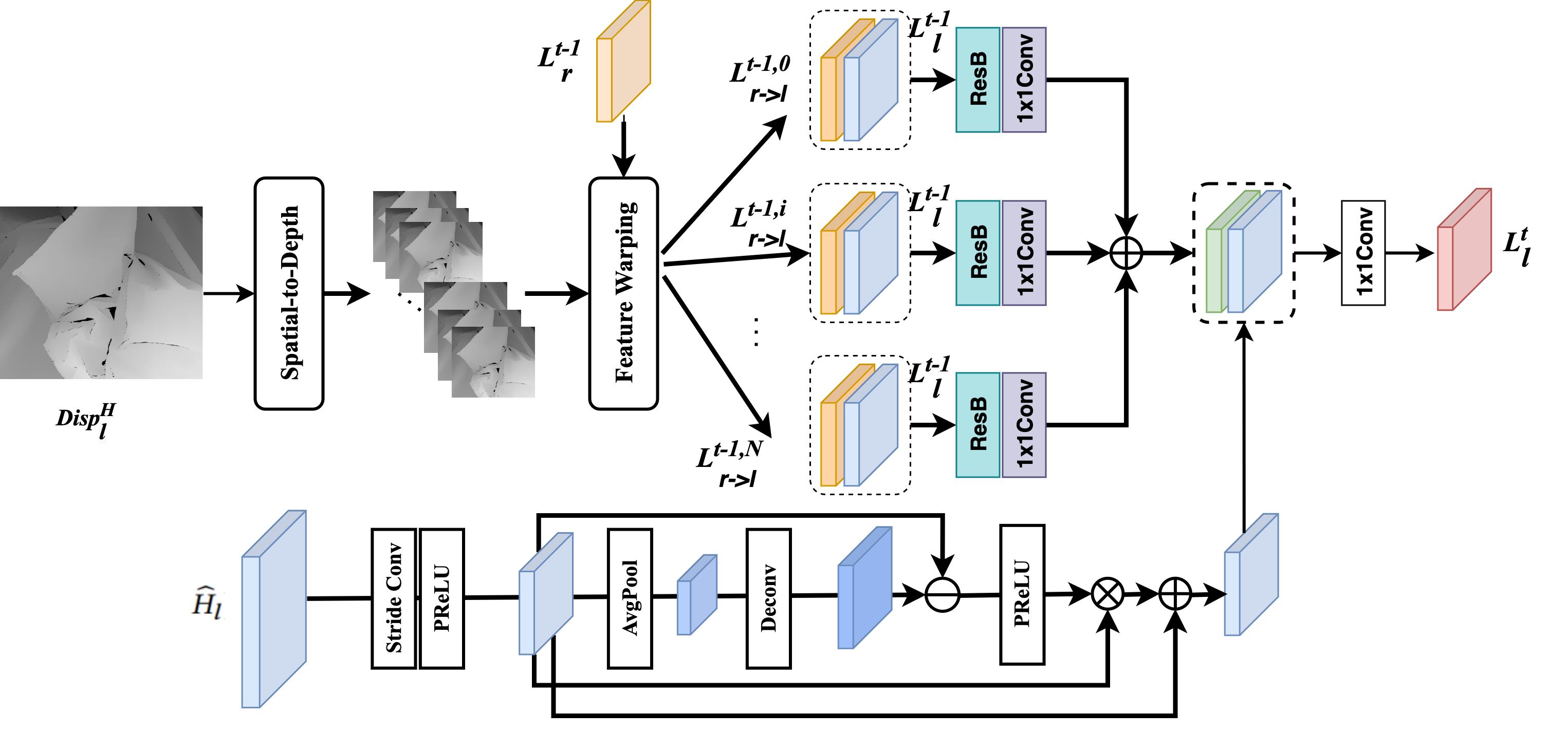}
  \begin{minipage}[c]{1\textwidth}
  \end{minipage}
  \caption{Illustration of our HR disparity information feedback (HRDIF) mechanism. (Please zoom in for details)}
  \label{fig-4}
  \vspace{-8pt}
\end{figure*}

\textbf{HR Disparity Estimation Module}
The downside to rely only on coarse matching is that the resulting correspondences lack fine details. Although LR correspondences have been demonstrated to benefit the stereo SR~\cite{Wang2019LearningPA, Song2020StereoscopicIS}, the low-level LR features limit the accuracy in correspondence matching, especially in high-frequency regions like object boundaries, which is the most important goal of SR. Thus, we suggest to also estimate the HR disparity map for more fine-gained correspondence information. To ensure the effectiveness of high-level HR features $H_{l}, H_{r}$, we connect the image reconstruction loss on the first step HR results $SR^{0}_{l}, SR^{0}_{r}$, thus the HR features $H_{l}, H_{r}$ can be seen as containing the information of HR images, and serve as reliable representations for HR disparity estimation. However, directly estimating from scratch costs massive computation cost, a more efficient strategy should be adopted. We found that the disparity maps $\mathbf{Disp}^{L}_{l}$ and $\mathbf{Disp}^{L}_{r}$ regressed from the parallax attention maps $\mathbf{M}_{l\rightarrow r}^{N}$ and $\mathbf{M}_{r\rightarrow l}^{N}$ have relative high accuracy in most regions (see the $1^{st}$ column of Tab.~\ref{disp1}), which can be obtained as:
\begin{equation}\label{eq-6}
\small
\setlength{\belowcaptionskip}{-12pt}
\begin{split}
\mathbf{Disp}^{L}_{l}=\sum_{k=0}^{W-1}k\times \mathbf{M}_{r\rightarrow l}(:,:,k),\\
\mathbf{Disp}^{L}_{r}=\sum_{k=0}^{W-1}k\times \mathbf{M}_{l\rightarrow r}(:,:,k),
 \end{split}
\end{equation}
where $W$ is the width of the input LR image. Thus, we only construct partial cost volumes $\mathbf{C}_{l}, \mathbf{C}_{r}$ based on coarse estimation and disparity residual hypotheses to achieve disparity maps with higher resolution and accuracy. As shown in Fig.\ref{fig-5}, the upsampled disparity maps ($up(\mathbf{Disp}^{L}_{l})$, $up(\mathbf{Disp}^{L}_{r})$) are used as initialization of the HR disparity estimation for the left and right view, respectively. The disparity searching range can then be narrowed, we task the network of only finding a residual to add or subtract from the coarse prediction, blending in high-frequency details.

Specially, we denote the disparity searching residual for each pixel in high resolution as $\Delta D$. Take the left view as an example, when performing $\times s$ SR, for the $m^{th}$ pixel in HR space, the disparity range for building the left cost volume is $[max(up(\mathbf{Disp}^{L}_{l})(m)-\Delta D/2,0), min(up(\mathbf{Disp}^{L}_{l})(m)+\Delta D/2, sW)]$. By uniformly sampling $P$ disparity hypotheses in this range (in this work, we set $P=\Delta D=24$), 3D cost volume with size $sH \times sW \times P$ can be obtained through feature correlation operation~\cite{Mayer2016ALD}. To learn more context information, we aggregate the cost volume using hourglass architecture. Then through soft-argmax operation, we can regress the HR disparity $\mathbf{Disp}^{H}_{l},\mathbf{Disp}^{H}_{r}$ for both view, with higher accuracy. For occlusion handling, we use the estimated disparity maps to check the geometric consistency and estimate the valid masks to be used in the loss functions:

\begin{equation}\label{eq-15}
\small
\setlength{\belowcaptionskip}{-12pt}
\begin{split}
\mathbf{V}_{l}=1-tanh(0.2\left | \mathbf{Disp}^{H}_{l}-Warp(\mathbf{Disp}^{H}_{r},\mathbf{Disp}^{H}_{l}) \right |),\\
\mathbf{V}_{r}=1-tanh(0.2\left | \mathbf{Disp}^{H}_{r}-Warp(\mathbf{Disp}^{H}_{l},\mathbf{Disp}^{H}_{r}) \right |),
 \end{split}
\end{equation}
where $Warp(\mathbf{Disp}^{H}_{r},\mathbf{Disp}^{H}_{l})$ represents using $\mathbf{Disp}^{H}_{l}$ to warp $\mathbf{Disp}^{H}_{r}$.

The HR disparity is in turn used to explore additional information from different views in the HR space, thus the registered HR features can be obtained by: $H_{r\rightarrow l}=Warp({H}_{r}, \mathbf{Disp}^{H}_{l})$, $H_{l\rightarrow r}=Warp({H}_{l}, \mathbf{Disp}^{H}_{r})$. For HR cross-view information aggregation, the residual-based module is adopted (similar to FFM), the only difference is that an additional attention map for each view is introduced to improve the aggregation reliability. Take the left view as example, the attention map measure the similarity of $H_{l}$ and $H_{r\rightarrow l}$: $Att_{l}=sigmoid(5f_{Conv1}(H_{l})\cdot f_{Conv2}(H_{r\rightarrow l}))$, where $f_{Conv1}$ and $f_{Conv2}$ are both $3 \times 3$ convolutional layers, $\cdot $ is the element-wise multiplication. Therefore, the aggregated HR left features $\widehat{H}_{l}$ are:

\begin{equation}\label{eq-16}
\small
\setlength{\belowcaptionskip}{-8pt}
\begin{split}
 Res_{l}&=(H_{r\rightarrow l}-H_{l})\cdot Att_{l}, \\
 \widehat{H}_{l}&=f_{CALayer}(f_{RDB}(Res_{l})+H_{l}). 
 \end{split}
\end{equation}
where $Att_{l}$ adaptively weights down the regions with too large difference with the original view and emphasis the regions that are favorable for providing complementary information. Similarly, we can get the aggregated right HR feature $\widehat{H}_{r}$. Afterwards, better SR images can be reconstructed through $\widehat{H}_{l}, \widehat{H}_{r}$:

\begin{equation}\label{eq-14}
\small
\setlength{\belowcaptionskip}{-8pt}
\begin{split}
 SR^{1}_{l}=f_{UP}(LR_{l})+f_{REC}(\widehat{H}_{l}), \\
 SR^{1}_{r}=f_{UP}(LR_{r})+f_{REC}(\widehat{H}_{r}). 
 \end{split}
\end{equation}

This section introduces a whole feed-forward pipeline for performing the two tasks. Three stages of task interactions have been shown: Firstly, LR disparity (correspondence) promotes image SR by adding extra details. Secondly, image SR promotes HR disparity estimation accuracy by providing fine-gained HR representations. Thirdly, the more accurate disparity promotes the quality of the SR images by aggregating features in the HR space. The interactions mentioned above all act in a straightforward way, however, we intend to further explore a more essential and intrinsic connection of the two tasks. 

\subsection{HRDIF Mechanism}\label{feedback}

The flow of information from the LR image to the final SR image is purely feed-forward in all previous stereo SR network architectures~\cite{Wang2019LearningPA, Wang2020ParallaxAF, Ying2020ASA, Song2020StereoscopicIS}, which cannot fully exploit effective high-resolution features in representing the LR to HR relation. The purely feed-forward network also makes it impossible for the HR disparity map to send useful information to the preceding low-level features, thus cannot refine these features in the SR process. To this end, we intend to project the useful information carried by the HR disparity back to preceding layers. Since the essential influence of the disparity to SR task is acting on the feature level, i.e., by registering the sup-pixel feature of two views and aggregating to obtain the enriched representations, we propose two strategies to feedback the HR disparity and act upon the feature space (Figure\ref{fig-4}, this illustration is based on the left view, the similar operation can be done on the right branch).

Firstly, the HR disparity information is embedded in the aggregated HR features $\widehat{H}_{l}, \widehat{H}_{r}$, thus we recommend to feed them back to refine the low-level features. Different from original feedback operation in ~\cite{Li2019FeedbackNF} that simply send the high-level features of the view back to low-level layer, our feedback HR features contain information both from intra-view and cross-view. To handle the spatial resolution gap, we back-project the HR features to LR space, and leverage a simple attention strategy to highlight the high-frequency regions in the downsampled features to compensate for the resolution loss. As shown in the downside branch of Fig.\ref{fig-4}, for the $t^{th}$ iteration, we first apply strided convolution to $\widehat{H}^{t-1}_{l}$ to obtain the back-projected feature $LB^{t}_{l}$. 

\begin{equation}\label{eq-7}
\small
\setlength{\belowcaptionskip}{-8pt}
LB^{t}_{l}=f_{DOWN}(\widehat{H}^{t-1}_{l}).
\end{equation}
Secondly, in order to get the high-frequency regions, we apply average pooling to $LB^{t}_{l}$, then a deconvolution layer is applied to project the feature back to original resolution, obtaining $\widetilde{LB^{t}_{l}}$. In addition, the attention map $W^{t}_{l}$ is calculated by computing the residual between $LB^{t}_{l}$ and $\widetilde{LB^{t}_{l}}$.

\begin{equation}\label{eq-8}
\small
\setlength{\belowcaptionskip}{-8pt}
\begin{split}
LP^{t}_{l}&=Avgpool(LB^{t}_{l}),\\
\widetilde{LB^{t}_{l}}&=f_{DeConv}(LP^{t}_{l}),\\
W^{t}_{l}&=PReLU(\widetilde{LB^{t}_{l}}-LB^{t}_{l}).
 \end{split}
\end{equation}
Then, the highlighted regions activated by $W^{t}_{l}$ is added to $LB^{t}_{l}$:
\begin{equation}\label{eq-9}
\small
\setlength{\belowcaptionskip}{-8pt}
LB^{t}_{l}=LB^{t}_{l}+\lambda (LB^{t}_{l}\cdot W^{t}_{l}),
\end{equation}
where $\lambda$ is a hyper-parameter used to control the importance of the attention weights. We name this feedback operation as AHFF (Aggregated HR Feature Feedback).

It is worth noting that one of the requirements that contains in a feedback system is providing an LR input at each iteration, i.e., to ensure the availability of low-level information which is needed to be refined. Thus, for the $t^{th}$ iteration, the LR feature $L^{t-1}_{l}$ from the $(t-1)^{th}$ iteration is meant to be refined by $LB^{t}_{l}$. Instead of directly leveraging the coarse original feature $L^{t-1}_{l}$, we propose the second HR disparity information feedback strategy to enrich the low-level representations. As shown in the upside of Figure.\ref{fig-4}, we first apply spatial-to-depth operation upon the $D^{H,t-1}_{l}\in \mathbb{R}^{sH \times sW}$, obtaining LR disparity cube of size $\mathbb{R}^{H\times W\times s^{2}}$. We leverage each disparity slice in the cube to warp $L^{t-1}_{r}$, obtaining $s^{2}$ warped feature maps of the right view. Each warped feature map is concatenated with the same left feature $L^{t-1}_{l}$, and each concatenated feature map is going through a residual block and a $1 \times 1$ convolution for fusion. Finally, we sum up the $s^{2}$ fused LR feature maps to get $\widehat{L}^{t-1}_{l}$. The operation can be defined as:
\begin{equation}\label{eq-10}
\small
\setlength{\belowcaptionskip}{-8pt}
\widehat{L}^{t-1}_{l}=\sum_{i=0}^{s^{2}}f_{fusion}(f_{ResB}(Concat(L^{t-1}_{l}, L^{t-1,i}_{r\rightarrow l}))).
\end{equation}
We name this strategy as LRE (Low-level Representations Enrichment).

Finally, $\widehat{L}^{t-1}_{l}$ and $LB^{t}_{l}$ are concatenated and fused to reduce the channel back to the same with ${L}^{t-1}_{l}$, and the new LR feature ${L}^{t}_{l}$ for the new iteration is generated according to:

\begin{equation}\label{eq-11}
\small
\setlength{\belowcaptionskip}{-8pt}
{L}^{t}_{l}=f_{fuse}(Concat(\widehat{L}^{t-1}_{l}, LB^{t}_{l})).
\end{equation}

In this way, the low-level features ${L}^{t}_{l}$ carry information from the HR disparity, and this feature enhancement dose favor to the whole pipeline right from the beginning. Finally, we adopt the last SR output as the final result.

\subsection{Loss Functions}\label{loss}
Since our work aims to achieve stereo SR and disparity estimation simultaneously, we set loss constraints for both tasks. Note that we learn the disparity in an unsupersived manner and do not use groundtruth (GT) disparities during the training phase. We introduce SR loss $\mathcal{L}_{SR}$, biPAM loss $\mathcal{L}_{BiPAM}$, and disparity loss $\mathcal{L}_{Disp}$ to train our network. The overall loss function of our network is defined as:
\begin{equation}
\small
\setlength{\belowcaptionskip}{-8pt}
\mathcal{L} = \mathcal{L}_{SR} + \lambda _{1}\mathcal{L}_{BiPAM} + \lambda _{2}\mathcal{L}_{Disp},
\end{equation}
where both $\lambda _{1}$ and $\lambda _{2}$ are set to 0.1 in this work.

\textbf{SR Loss.}
The SR loss is essentially an $L_{1}$ loss function that is used to measure the difference between the SR images and GT images, i.e., for T iterations, 
\begin{equation}
\small
\setlength{\belowcaptionskip}{-8pt}
\begin{split}
\mathcal{L}_{SR} &= \sum_{t=0}^{T} \parallel \mathbf{SR}_{l}^{t,0} - \mathbf{HR}_{l} \parallel_1 +  \parallel \mathbf{SR}_{r}^{t,0} - \mathbf{HR}_{r} \parallel_{1} \\
&+ \parallel \mathbf{SR}_{l}^{t,1} - \mathbf{HR}_{l} \parallel_1 + \parallel \mathbf{SR}_{r}^{t,1} - \mathbf{HR}_{r} \parallel_{1},
\end{split}
\end{equation}
where $\mathbf{SR}_{l}$ and $\mathbf{SR}_{r}$ represent the restored left and right images, and $\mathbf{HR}_{l}$ and $\mathbf{HR}_{r}$ represent their corresponding HR images.

\textbf{BiPAM Loss.}
We formulate the BiPAM loss as a combination of photometric, smoothness, cycle and consistency terms, connecting to bi-directional parallax-attention maps $\mathbf{M}^{t}_{r\rightarrow l}$, $\mathbf{M}^{t}_{l\rightarrow r}$, t=1,...,T. That is, $\mathcal{L}_{BiPAM}=\mathcal{L}_{photo} + \mathcal{L}_{cycle} + \mathcal{L}_{smooth} + \mathcal{L}_{cons}$. The loss is employed in a residual manner~\cite{Wang2020SymmetricPA} to overcome illuminance variation. Please refer to \cite{Wang2020SymmetricPA} for details.

\textbf{Disparity Loss.}
Besides tying loss on the parallax-attention maps, we also enforce direct constraints on all the estimated disparity maps, namely $D^{L,t}_{l}$, $D^{L,t}_{r}$, $D^{H,t}_{l}$, $D^{H,t}_{r}$ for $t=1,...,T$. We first penalize the reconstruction loss on HR images using each disparity map (LR disparity upsamples to the same size of HR images), for the left view, 
\begin{equation}
\small
\setlength{\belowcaptionskip}{-8pt}
\begin{split}
\mathcal{L}_{rc}^{l}&=\frac{1}{N}\sum_{p\in \mathbf{V}^{t}_{l},t=1}^{t=T}\alpha \frac{1-\mathcal{S}(\mathbf{HR}_{l}(p),\mathbf{HR}^{t}_{r\rightarrow l}(p))}{2}\\
&+(1-\alpha )\left \| \mathbf{HR}_{l}(p)- \mathbf{HR}^{t}_{r\rightarrow l}(p)\right \|_{1}, t=1,...,T,
\end{split}
\end{equation}
where $\mathbf{HR}^{t}_{r\rightarrow l}=Warp(\mathbf{HR}_{r},\mathbf{Disp}^{H,t}_{l})$. $\mathcal{S}$ is a structural similarity index (SSIM) function, $p$ represents a valid pixel in the valid mask, $N$ is the number of valid pixels, and $\alpha$ is empirically set to 0.85. The loss for the right view is also calculated as the similar method.

Moreover, we constrain edge-aware smoothness loss on HR disparity, which is defined as: 
\begin{equation}
\small
\setlength{\belowcaptionskip}{-8pt}
\begin{split}
\mathcal{L}_{s}^{l}&=\frac{1}{N}\sum_{\mathbf{p}\in \mathbf{V}_{l}}^{}\left \| \triangledown _{x}\mathbf{D}_{l}^{HR,t}(\mathbf{p}) \right \|_{1}e^{-\left \| \triangledown _{x}\mathbf{HR}_{l}(\mathbf{p}) \right \|_{1}}
\\
&+\left \| \triangledown _{y}\mathbf{D}_{l}^{HR,t}(\mathbf{p}) \right \|_{1}e^{-\left \| \triangledown _{y}\mathbf{HR}_{l}(\mathbf{p}) \right \|_{1}}, t=1,...,T,
\end{split}
\end{equation}
where $\triangledown _{x}$ and $\triangledown _{y}$ are gradients in the $x$ and $y$ directions respectively.

Finally, residual based cycle and consistency losses~\cite{Wang2020SymmetricPA} are also used to constrain HR disparity maps. The total disparity loss can be written as: $\mathcal{L}_{Disp}=\mathcal{L}_{rc} + \mathcal{L}^{HR}_{cycle} + \mathcal{L}^{HR}_{cons} + 0.1*\mathcal{L}_{s}$.

\begin{table*}[!t]
\centering
\caption{Quantitative results achieved by different methods on the KITTI 2012, KITTI 2015, Middlebury, and Flickr1024 datasets. $\#P$ represents the number of parameters of the networks. Here, PSNR$/$SSIM values achieved on both the left images (i.e., \textit{Left}) and a pair of stereo images (i.e., $\left(\textit{Left}+\textit{Right}\right)/2$) are reported. The best results are in \textbf{bold faces} and the second best results are \underline{underlined}.} \label{TabQuantitative}
\renewcommand\arraystretch{1}
\vspace{-10px}
\resizebox{\textwidth}{!}
{
\begin{tabular}{lccccccccc}
\toprule
\multirow{2}*{Method} & \multirow{2}*{Scale} & \multirow{2}*{$\#P$} & \multicolumn{3}{c}{\textit{Left}} & \multicolumn{4}{c}{$\left(\textit{Left}+\textit{Right}\right)/2$}\\
\cmidrule(lr){4-6} \cmidrule(lr){7-10}
         &      &           & KITTI 2012 & KITTI 2015 & Middlebury & KITTI 2012 & KITTI 2015 & Middlebury & Flickr1024\\
\hline
VDSR & $\times$2 & 0.66M & 30.17$/$0.9062 & 28.99$/$0.9038 & 32.66$/$0.9101 & 30.30$/$0.9089 & 29.78$/$0.9150& 32.77$/$0.9102 & 25.60$/$0.8534\\
EDSR & $\times$2 & 38.6M & 30.83$/$0.9199 & 29.94$/$0.9231 & 34.84$/$\underline{0.9489} &30.96$/$0.9228 & 30.73$/$0.9335 & \underline{34.95}$/$\underline{0.9492} & \underline{28.66}$/$0.9087 \\
RDN & $\times$2 & 22.0M  & 30.81$/$0.9197 & 29.91$/$0.9224 & \underline{34.85}$/$0.9488 &30.94$/$0.9227 & 30.70$/$0.9330 & 34.94$/$0.9491 & 28.64$/$0.9084 \\
RCAN & $\times$2 & 15.3M & 30.88$/$0.9202 & 29.97$/$0.9231 & 34.80$/$0.9482 & 31.02$/$0.9232 & 30.77$/$0.9336 & 34.90$/$0.9486 & 28.63$/$0.9082 \\
StereoSR & $\times$2 &1.08M & 29.42$/$0.9040 & 28.53$/$0.9038 & 33.15$/$0.9343 & 29.51$/$0.9073 & 29.33$/$0.9168 & 33.23$/$0.9348 & 25.96$/$0.8599 \\
PASSRnet & $\times$2 & 1.37M & 30.68$/$0.9159 & 29.81$/$0.9191 & 34.13$/$0.9421 & 30.81$/$0.9190 & 30.60$/$0.9300 & 34.23$/$0.9422 & 28.38$/$0.9038 \\
IMSSRnet & $\times$2 & 6.84M & 30.90$/$- & 29.97$/$- & 34.66$/$- & 30.92$/$- & 30.66$/$- & 34.67$/$- & -$/$- \\
iPASSR & $\times$2 & 1.37M & \underline{30.97}$/$\underline{0.9210} & \underline{30.01}$/$\underline{0.9234} & 34.41$/$0.9454 & \underline{31.11}$/$\underline{0.9240} & \underline{30.81}$/$\underline{0.9340} & 34.51$/$0.9454 & 28.60$/$\underline{0.9097} \\
SSRDE-FNet (ours)   & $\times$2 & 2.10M & \textbf{31.08}$/$\textbf{0.9224} & \textbf{30.10}$/$\textbf{0.9245} & \textbf{35.02}$/$\textbf{0.9508} & \textbf{31.23}$/$\textbf{0.9254} & \textbf{30.90}$/$\textbf{0.9352} & \textbf{35.09}$/$\textbf{0.9511} & \textbf{28.85}$/$\textbf{0.9132} \\
\hline
VDSR &  $\times$4 & 0.66M & 25.54$/$0.7662 & 24.68$/$0.7456 & 27.60$/$0.7933 & 25.60$/$0.7722 & 25.32$/$0.7703 & 27.69$/$0.7941 & 22.46$/$0.6718 \\
EDSR &  $\times$4 & 38.9M & 26.26$/$0.7954 & 25.38$/$0.7811 & 29.15$/$\underline{0.8383} & 26.35$/$0.8015 & 26.04$/$0.8039 & 29.23$/$0.8397 & 23.46$/$0.7285 \\
RDN &  $\times$4 & 22.0M  & 26.23$/$0.7952 & 25.37$/$0.7813 & 29.15$/$0.8387 & 26.32$/$0.8014 & 26.04$/$0.8043 & 29.27$/$\underline{0.8404} & 23.47$/$\underline{0.7295} \\
RCAN &  $\times$4 & 15.4M & 26.36$/$0.7968 & 25.53$/$0.7836 & \underline{29.20}$/$0.8381 & 26.44$/$0.8029 & 26.22$/$0.8068 & \underline{29.30}$/$0.8397 & \underline{23.48}$/$0.7286 \\
StereoSR  &  $\times$4 & 1.42M   & 24.49$/$0.7502 & 23.67$/$0.7273 &27.70$/$0.8036 & 24.53$/$0.7555 & 24.21$/$0.7511 & 27.64$/$0.8022 & 21.70$/$0.6460 \\
PASSRnet  &  $\times$4 & 1.42M   & 26.26$/$0.7919 & 25.41$/$0.7772 &28.61$/$0.8232 & 26.34$/$0.7981 & 26.08$/$0.8002 & 28.72$/$0.8236 & 23.31$/$0.7195 \\
SRRes+SAM  &  $\times$4 & 1.73M  & 26.35$/$0.7957 & 25.55$/$0.7825 & 28.76$/$0.8287 & 26.44$/$0.8018 & 26.22$/$0.8054 & 28.83$/$0.8290 & 23.27$/$0.7233 \\
IMSSRnet &  $\times$4 & 6.89M  & 26.44$/$- & 25.59$/$- & 29.02$/$- & 26.43$/$- & 26.20$/$- & 29.02$/$- & -$/$- \\
iPASSR  &  $\times$4 & 1.42M  & \underline{26.47}$/$\underline{0.7993} & \underline{25.61}$/$\underline{0.7850} & 29.07$/$0.8363 & \underline{26.56}$/$\underline{0.8053} & \underline{26.32}$/$\underline{0.8084} & 29.16$/$0.8367 & 23.44$/$0.7287 \\
SSRDE-FNet (ours)  & $\times$4 & 2.24M  & \textbf{26.61}$/$\textbf{0.8028} & \textbf{25.74}$/$\textbf{0.7884} & \textbf{29.29}$/$\textbf{0.8407} & \textbf{26.70}$/$\textbf{0.8082} & \textbf{26.43}$/$\textbf{0.8118} & \textbf{29.38}$/$\textbf{0.8411} & \textbf{23.59}$/$\textbf{0.7352} \\
\bottomrule
\end{tabular}}
\vspace{-8px}
\end{table*}

\section{Experiments}
\subsection{Experimental Settings}
Following iPASSR\cite{Wang2020SymmetricPA}, we adopt 60 Middlebury images and 800 images from Flickr1024~\cite{Wang2019Flickr1024AL} as the training dataset during training. For images from the Middlebury dataset, we followed~\cite{Jeon2018EnhancingTS,Wang2019LearningPA,Wang2020ParallaxAF,Ying2020ASA, Wang2020SymmetricPA} to perform bicubic downsampling by a factor of 2 to generate HR ground truth images to match the spatial resolution of Flickr1024 dataset. To produce LR images, we downscale the HR images on particular scaling factors by using the bicubic operation and then cropped $30\times90$ patches with a stride of 20 as input samples. Our network was implemented using PyTorch and trained on NVIDIA V100 GPU. All models were optimized by the Adam~\cite{Kingma2015AdamAM} with $\beta _{1}=0.9$ and $\beta _{2}=0.999$. The batch size is set to $16$, the initial learning rate is set to $2\times10^{-4}$ and reduced to half after every 30 epochs. 

To evaluate SR results, 20 images from KITTI 2012\cite{Geiger2012AreWR}, 20 images from KITTI 2015\cite{Menze2015ObjectSF}, 5 images from Middlebury, and 112 images from Flickr1024 are utilized as the test dataset. For fair comparison with~\cite{Jeon2018EnhancingTS,Wang2019LearningPA,Ying2020ASA}, we followed these methods to calculate peak signal-to-noise ratio (PSNR) and structural similarity (SSIM) scores on the left views with their left boundaries (64 pixels) being cropped, and these metrics were calculated on RGB color space. Moreover, to comprehensively evaluate the quality of the reconstructed stereo SR image, we also report the average PSNR and SSIM scores on stereo image pairs (i.e., $\left(\textit{Left}+\textit{Right}\right)/2$) without any boundary cropping. Meanwhile, in order to evaluate disparity estimation results, we apply the end-point-error (EPE) in both non-occluded region (NOC) and all (ALL) pixels.

\begin{figure*}[t]
  \centering
  \includegraphics[width=17cm]{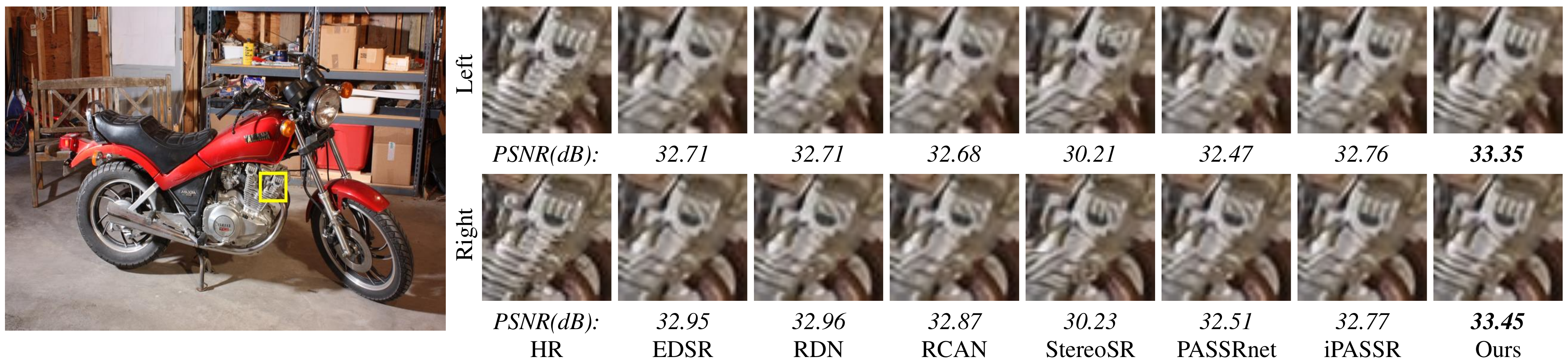}
  \vspace{-10px}
  \caption{Qualitative results (×2) on image “motorcycle” from Middlebury dataset.}
  \label{fig-4}
  \vspace{-12pt}
\end{figure*}

\begin{figure*}[t]
  \centering
  \includegraphics[width=17cm]{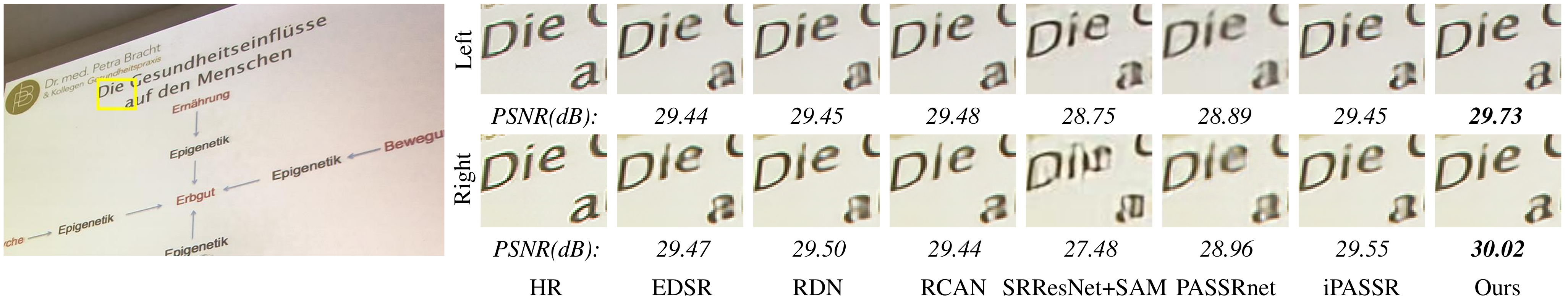}
  \vspace{-10px}
  \caption{Qualitative results (×4) on image “testing 2” from Flickr1024 dataset.}
  \label{fig-5}
  \vspace{-8pt}
\end{figure*}

\subsection{Comparisons with SOTA Methods}
We compare SSRDE-FNet with several state-of-the-art methods, including four SISR methods(VDSR, EDSR, RDN, and RCAN) and five stereo image SR methods (i.e., StereoSR, PASSRnet, SRResNet+SAM, IMSSRnet, and iPASSR). Moreover, to achieve fair comparison with SISR methods, we retrained these methods on the same training datasets as our method.

\textbf{Quantitative Evaluations:}
In Table~\ref{TabQuantitative}, we show the
quantitative comparisons with these SR methods. Among both SISR and stereo image SR methods, our FSSRHD-net achieves the best results on all datasets and upsampling factors ($\times 2$, $\times 4$). This fully demonstrates the effectiveness and advancement of the proposed SSRDE-FNet.

\textbf{Visual Comparison:}
In Figures~\ref{fig-4} and ~\ref{fig-5}, we show the visual comparisons on $\times 2$ and $\times 4$, respectively. According to the figure, we can clearly observe that
most compared SR methods cannot recover clear and right image edges. In contrast, our SSRDE-FNet can reconstruct high-quality SR images with rich details and clear edges. This further validates the effectiveness of our SSRDE-FNet.

\subsection{Ablation Study}
In order to verify the effectiveness of the proposed mutually boost strategies, we designed a series of ablation experiments. In addition, all ablation studies are conducted on the $\times4$ stereo image SR task. It is worth noting that the baseline model does not use the HR disparity estimation mechanism and the feedback strategy. This means that the baseline model has only one step of SR reconstruction, as shown in Figure\ref{fig-3}. 

\textbf{Effectiveness of HR disparity estimation boost SR}

1)Effectiveness of the HR disparity estimation method. In order to verify that the feature aggregation by the HR disparity in HR space benefits the SR performance, we designed three models, including "baseline", "baseline+ Up disp", and "baseline + HR disp". Among them, "baseline+ Up disp" means that the high-resolution disparity directly achieved by the interpolation operation and "baseline+ HR disp" represents our proposed method. Meanwhile, all of these three model are in purely feed-forward manner. The PSNR and SSIM results are presented in Table~\ref{component}. According to these results, we can draw the following conclusions: (1). high-resolution disparity can effectively improve the quality of the reconstructed SR images; (2). the more precise disparity can bring higher performance improvement; (3) the high-resolution disparity provided by our method enables the model to achieve the best results.

\begin{table}
\caption{Ablation study on different settings of SSRDE-FNet on Middlebury. The average PSNR and SSIM score of the SR left and right images are shown.}
\label{component}
\renewcommand\arraystretch{0.9}
\vspace{-10pt}
    \small
    \setlength{\tabcolsep}{0.7mm}
	\centering
	\begin{tabular}{lcccccc}
	    \toprule
	   \multirow{2}{*}{Method} & \multicolumn{2}{c}{Disparity method} & \multicolumn{2}{c}{HRDIF} & \multirow{2}{*}{HFF} & \multirow{2}{*}{PSNR/SSIM}\\
	   \cmidrule(lr){2-3} \cmidrule(lr){4-5}
		& {Up disp} & {HR disp} & AHFF & LRE & &  \\
		\midrule
		 baseline &  &  &  & & & 29.16/0.8361 \\
		 baseline + Up disp & $\checkmark$ & & & &  & 29.20/0.8370 \\
		 baseline + HR disp & & $\checkmark$& & & & 29.27/0.8383 \\
		 SSR-FNet &  & & & & $\checkmark$ & 29.27/0.8385 \\
		 SSRDE-FNet w/o LRE &  & $\checkmark$ & $\checkmark$ & & & 29.35/0.8407 \\
		 SSRDE-FNet (Ours) &  & $\checkmark$ & $\checkmark$ & $\checkmark$ & & \textbf{29.38/0.8411} \\
		\bottomrule
	\end{tabular}
\vspace{-8px}
\end{table}

2) Effectiveness of the HR disparity information feedback mechanism (HRDIF): To verify that the HR disparity truly contribute to stereo SR in the HRDIF mechanism, but not just the original feedback operation that plays a major role, we compare two models that both have the feedback operation. The variant removes the HR disparity estimation model, directly use the $H_{l}$ and $H_{r}$ as the high-level features to feedback. We name this variant as \textbf{SSR-FNet} (Stereo SR Feedback Network), which also means adding HR Feature Feedback (HFF) to the baseline.  The feedback manner in the variant is just concatenating the down-projected HR feature and the low-level features of the last iteration. Although noticeable improvement can be observed, the PSNR drops 0.11 dB as compared to our SSRDE-FNet. The experiment indicates that our method does benefit from the HR disparity information feedback mechanism, instead of only rely on the power of the original feedback structure. Moreover, to verify the effectiveness of strategy of the low-level representations enhancement (LRE) in HRDIF, we remove this operation and directly concatenate $L^{t-1}_{l}$ and $LB^{t}_{l}$ for the $t^{th}$ iteration, a slight PSNR drop can be observed.

3) SR performance improvements in a single inference: As mentioned, each iteration of SSRDE-FNet contains two SR reconstruction steps. In our experiments, we iterate the network twice (T=2) to balance the efficiency and performance.  We then compare the PSNR values of all intermediate SR images. The results are shown in Table~\ref{SRchange}. Each intermediate result outperforms the former one, and the final result achieves a PSNR gain of 0.22dB over the first result. This demonstrates that the HR disparity surely benefits the information flow across time.

\begin{table}
\caption{The PSNR changing of intermediate SR outputs on Middlebury.}
\label{SRchange}
\vspace{-10pt}
\renewcommand\arraystretch{0.9}
    \small
	\centering
	\setlength{\tabcolsep}{3.6mm}{
	\begin{tabular}{lcccc}
	    \toprule
	   & \multicolumn{2}{c}{Iteration 1} &  \multicolumn{2}{c}{Iteration 2}\\
	    \cmidrule(lr){2-3} \cmidrule(lr){4-5}
		& {Step 1} & {Step 2} & Step 1 & Step 2 \\
		\midrule
		 Middlebury & 29.16 & 29.25 & 29.32 & 29.38 \\
		\bottomrule
	\end{tabular}}
	\vspace{-16px}
\end{table}

\textbf{Effectiveness of SR boost disparity estimation}

	\begin{table}
		\caption{\textcolor{black}{Average disparity EPE errors (lower is better) on KITTI 2012 and KITTI 2015 for $4\times$ SR. Best results are shown in boldface.}}
		\label{disp1}
		\vspace{-10pt}
		\renewcommand\arraystretch{0.9}
		\small
		\begin{center}
			\small
			\setlength{\tabcolsep}{1mm}{
				\begin{tabular}{lccc|cc}
					\toprule
					\multicolumn{2}{c}{} 
					& \tabincell{c}{Baseline\\disparity} 
					& \tabincell{c}{Estimated HR\\disparity} 
					& \tabincell{c}{\textcolor{black}{PASSRnet}\\ \cite{Wang2019LearningPA}}
					& \tabincell{c}{\textcolor{black}{iPASSR}\\ \cite{Wang2020SymmetricPA}}
					\tabularnewline
					\cline{1-6}
					\hline
					\multirow{2}{*}{KITTI 2012}& Noc 
					& 6.72 & \textbf{3.90} & 11.33 & 7.88
					\tabularnewline
					& All & 7.81 & \textbf{5.12} & 12.29 & 8.96
					\tabularnewline
					\hline
					\multirow{2}{*}{KITTI 2015}& Noc 
					& 5.71 & \textbf{3.52} & 9.36 & 6.57
					\tabularnewline
					& All &  6.38 & \textbf{4.28} & 9.91 & 7.20
					\tabularnewline
					\bottomrule
					
			\end{tabular}}
		\end{center}
		\vspace{-8px}
	\end{table}
	
\begin{figure}[t]
  \centering
  \includegraphics[width=8.5cm]{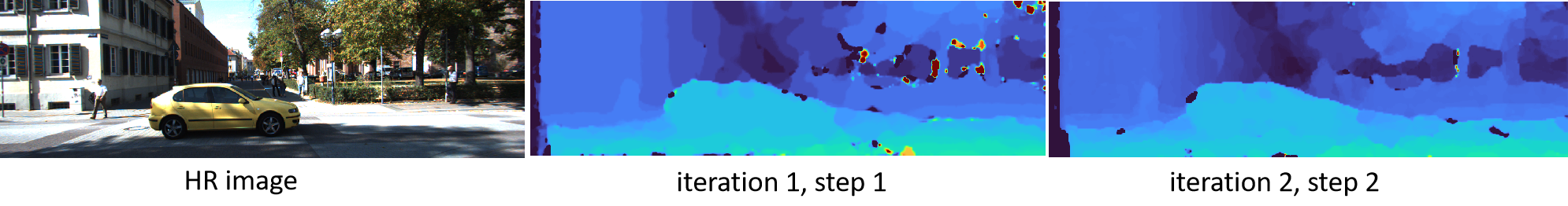}
  \vspace{-12px}
  \caption{Visualization result of the disparity map on KITTI 2015.}
  \label{fig-6}
  \vspace{-12pt}
\end{figure}

1) Comparison of disparity accuracy: 
We compare the estimated HR disparity and upsampled disparity of the baseline to the ground truth on the KITTI2012 and KITTI2015 datasets, shown in Table.\ref{disp1}. We also include the disparity regressed from two stereo SR methods for comparison, including PASSRnet and iPASSR. The disparity maps estimated from LR stereo images are upsampled for fair evaluation. Even using our baseline, our disparity EPE error is obviously lower than that of other state-of-the-art stereo SR methods. By interacting stereo SR task and disparity estimation task in our network, the final HR disparity become much more accurate as compared to the straightforward baseline, with about $2\sim 3$ pixel EPE error drop. A visualization disparity result is shown in Figure.\ref{fig-6}.

\begin{table}
\caption{Disparity accuracy improvements across inference time on KITTI 2012 and KITTI 2015 dataset.}
\label{dispchange}
\vspace{-8pt}
    \renewcommand\arraystretch{0.9}
    \small
	\centering
	\setlength{\tabcolsep}{2mm}{
	\begin{tabular}{lccccc}
	    \toprule
	   & & \multicolumn{2}{c}{Iteration 1} &  \multicolumn{2}{c}{Iteration 2}\\
	    \cmidrule(lr){3-4} \cmidrule(lr){5-6}
		& & {Step 1} & {Step 2} & Step 1 & Step 2 \\
		\midrule
		 \multirow{2}{*}{KITTI 2012}& Noc & 7.13 & 6.50 & 4.59 & 3.90 \\
		 & ALL & 8.14 & 7.53 & 5.79 & 5.12\\
		 \hline
		 \multirow{2}{*}{KITTI 2015}& Noc & 6.98 & 6.47 & 4.06 & 3.52 \\
		 & ALL & 7.60 & 7.11 & 4.81 & 4.28
		 \\ 
		\bottomrule
	\end{tabular}}
	\vspace{-8px}
\end{table}

2) The disparity accuracy improvements within a single inference of SSRDE-FNet: To show the changing process of the disparity estimation accuracy, we calculate the EPE error on each intermediate disparity estimation in a single inference process of SSRDE-FNet. The mean EPE error change in KITTI 2012 and KITTI 2015 are shown in Tab.~\ref{dispchange}. It can be observed that in each iteration, the estimated HR disparity (step2) has $0.5\sim 0.6$ pixel EPE error drop compared to the coarse estimation (step1). More obvious disparity accuracy improvements can be achieved after the HRDIF, since the low-level features are refined and lead to better disparity accuracy right from the LR space. The results above demonstrate that both stereo SR and disparity estimation are improved along time. 

\section{Conclusion}
In this work, we propose to explore the mutually boosted property of stereo image super-resolution and high-resolution disparity estimation, and build a novel end-to-end deep learning framework, namely SSRDE-FNet. Our model is essentially a feedback network with a proposed HR Disparity Information Feedback (HRDIF) mechanism. By fully interacting the two tasks and making guidance to each other, we achieve to improve both tasks during a single inference. Experiments have demonstrated our state-of-the-art stereo SR performance and the disparity estimation improvements.


\bibliographystyle{ACM-Reference-Format}
\bibliography{sample-sigconf}

\end{document}